\documentclass[lettersize,kjournal]{IEEEtran}
\IEEEoverridecommandlockouts

\usepackage{cite}
\usepackage{float}
\usepackage{stfloats}
\usepackage{fancyhdr}

\usepackage{bbding}

\usepackage{setspace}

\usepackage{color}
\usepackage{amsmath}
\usepackage{graphicx}
\graphicspath{{./figures/}}
\usepackage{algorithm}
\usepackage{algorithmic}
\usepackage{soul, color, xcolor}

\usepackage{epsfig}
\usepackage{epstopdf}

\usepackage{subfigure}

\usepackage{bm}

\usepackage{setspace}
\usepackage[colorlinks,linkcolor=blue,citecolor=blue]{hyperref}
\usepackage{orcidlink}

\usepackage{array}

\usepackage{textcomp}
\usepackage{url}
\usepackage{verbatim}
\usepackage{balance}

\usepackage{tabularx}
\usepackage{booktabs}
\usepackage{makecell}
\usepackage{booktabs}

\usepackage[capitalise, nameinlink]{cleveref}  

\begin{document}
	\title{6G Communication Networks Enabling Embodied Agents: Architecture and Prototype}

	\author{\IEEEauthorblockN{
		    Lipeng Dai $^{\orcidlink{0000-0002-1924-262X}}$,
			Luping Xiang  $^{\orcidlink{0000-0003-1465-6708}}$, \IEEEmembership{Senior~Member, IEEE},
			and Kun Yang, \IEEEmembership{Fellow, IEEE} 
			}
			\vspace{-0.5 cm}\\

 			\thanks{
			Lipeng Dai, Luping Xiang, and Kun Yang are with the State Key Laboratory of Novel Software Technology, Nanjing University, Nanjing, 210008, China, Institute of Intelligent Networks and Communications (NINE), Nanjing University (Suzhou Campus), Suzhou, 215163, China, email:  dlp1022@163.com, luping.xiang@nju.edu.cn, kunyang@nju.edu.cn. (\textit{Corresponding Author: Luping Xiang.})	
			}
	}
	\maketitle
	
\begin{abstract}
Embodied agents, which couple intelligent decision-making with physical actuation in the real world, impose far more stringent and heterogeneous communication requirements than purely software-based agents. 
While 6G promises sub-millisecond latency, ultra-high reliability, native intelligence, and integrated sensing, systematic studies on how to exploit these capabilities for embodied agent communication remain limited. 
This article investigates 6G-enabled communication systems for embodied agents from both conceptual and engineering perspectives. 
First, we review the concept, embodiment value of embodied agents, and clarify their distinctions from disembodied agents. 
Then, we analyse the symbiotic relationship between embodied agents and 6G networks.
We highlight how key 6G enablers can support the stringent requirements of human-robot interaction.
Furthermore, we demonstrate the proactive role of embodied agents in bolstering communication networks through coverage extension, environmental sensing, and physical world understanding.
Building on these insights, we propose a hierarchical communication architecture for human-robot remote interaction, comprising a human-intent perception layer, an open radio access network (O-RAN)-based transport layer, an intelligent intermediary layer, and an embodiment layer. 
To validate its feasibility, we implement an end-to-end prototype that integrates a haptic device, an industrial robotic arm, an intermediary platform, and a 5G O-RAN testbed. 
Experimental results demonstrate millisecond-level latency and stable closed-loop operation, confirming the practicality of the proposed architecture and providing a reference for future 6G-embodied agent research and industrial deployments.
\end{abstract}
	

	
\section{Introduction}
\label{Introduction}
\IEEEPARstart{C}{ommunication} networks have traditionally been architected around client-server models, with traffic patterns dominated by north-south flows.
In recent years, however, the explosive proliferation of connected devices and the diversification of service types have resulted in a substantial rise in east-west traffic, namely lateral data exchange among endpoints and between endpoints and edge nodes, exemplified by device state synchronisation and the dissemination of collaborative control commands.
Concurrently, rapid advances in artificial intelligence (AI) are driving the emergence of a novel communication paradigm organised around intelligent agents \cite{Agentic_AI_6G_JSAC, Large_AI_Models_6G_COMST}.
The cooperative communication and data exchange among such agents are expected to become the principal contributor to future east-west traffic growth, while simultaneously accelerating the transition of communication paradigms from human-centric to agent-centric designs.

Within this paradigm shift, future societies are anticipated to be populated by a large number of intelligent agents that directly interact with the physical environment, i.e., embodied agents.
In an era characterised by embodied intelligence, a wide spectrum of transformative applications and products will be enabled, including intelligent manufacturing systems where robotic arms autonomously execute component processing, smart homes in which service robots perform cleaning, caregiving, and integrated appliance management, and advanced healthcare scenarios supporting remote surgical guidance via networked diagnostic equipment.
The large-scale deployment and frequent coordination of embodied agents will further amplify east-west traffic volumes and, in parallel, impose stringent performance requirements on communication networks, such as ultra-low latency and high reliability.
The evolution toward 6G systems offers a critical opportunity to realise the intelligent interconnection of embodied agents.
Featuring sub-millisecond latency, ultra-high reliability, and built-in intelligence, 6G exhibits technical characteristics that are highly congruent with the operational demands of embodied agents.

Despite this strong technological affinity between embodied agents and 6G, comprehensive investigations into their deep, system-level integration remain limited.
Existing studies largely concentrate on the collaboration between disembodied agents and 6G, with an emphasis on interaction mechanisms and network architecture design within fully digital environments \cite{6G_Agent_Architecture_CSM,6G_multiAgent_LLM_WCM,6G_LLM_WCM}.
Although \cite{6G_AgentNet_CM} introduces the AgentNet framework, which can in principle accommodate embodied agent access, its primary focus lies in the architectural aspects of the network rather than in exploring holistic integration between embodied agents and 6G.
Moreover, while \cite{6G_Embodiment_WCM} offers an insightful discussion of the fundamental relationship between embodied agents and 6G, its analysis is largely prospective in nature, asserting that their convergence is an inevitable pathway toward artificial general intelligence (AGI) without addressing the key technical challenges associated with enabling 6G to support embodied agent operation.
A detailed comparative analysis of related work is presented in \cref{Comparison of Research}.

\begin{table*}[!t]
    \centering
    \caption{Comparison of State-of-the-Art Research on 6G-Agent Systems}
    \label{Comparison of Research} 
    \begin{tabular}{ m{1.2cm}<{\centering} m{2cm}<{\centering}  m{13cm} } 
        \toprule
		Literature & Agent Type & Focus  \\
		\toprule
        \cite{6G_Agent_Architecture_CSM}   & Disembodied & Establishing an intelligent unified network architecture aligned with 6G core metrics. \\
        \midrule
		\cite{6G_multiAgent_LLM_WCM}  & Disembodied & Exploring solutions for the appropriate application of LLMs in 6G, and proposing a multi-agent system named CommLLM.\\
        \midrule
		\cite{6G_LLM_WCM}   & Disembodied & Exploring the challenges of constructing LLM agents in 6G mobile edge computing environments, and proposing a split learning system that leverages the collaboration between mobile devices and edge servers. \\
		\midrule
		\cite{6G_AgentNet_CM}  & Disembodied \& embodied & Exploring the challenges and requirements for constructing agent-based intelligent networks, and proposing a framework named AgentNet to support communication and networking among agents, including embodied agents. \\
		\midrule
		\cite{6G_Embodiment_WCM}  & Embodied & Exploring the integration of 6G and embodied agents as a pathway towards achieving AGI. \\
		\midrule
		This article  & Embodied & Analyzing embodied agents and their interplay with 6G, and proposing a communication architecture for human-robot remote collaboration and its prototype implementation. \\
		\bottomrule
    \end{tabular}
\end{table*}

Given that embodied agents are realised as physical entities that directly interact with their surrounding environment, their communication requirements deviate substantially from those of purely software-based agents, thereby introducing a series of new technical challenges for future intelligent interconnection scenarios.
To fully unlock the potential of hardware-based agents and advance the vision of seamless interoperability between disembodied and embodied agents, this article conducts specialised research on communication systems for embodied agents. 
In particular, we first review the definition and key characteristics of embodied agents, then analyse their symbiotic enabling relationship with 6G networks, and subsequently propose a communication system architecture to support human–robot remote interaction.
The practicality of the proposed architecture is demonstrated through the development and implementation of a prototype system, which offers meaningful guidance for subsequent academic studies and industrial-scale engineering deployments.

\section{What Are Embodied Agents?}
\label{What Are Embodied Agents}

\subsection{Brief History}
\label{Brief History}

Since Alan Turing first touched upon the concept of embodied intelligence in his seminal 1950 paper \emph{Computing Machinery and Intelligence}, exploring the interaction between intelligence and the physical world has emerged as an enduring undercurrent throughout AI research.
Claude Shannon created the \emph{Theseus} mouse in 1950, widely acknowledged as one of the earliest analogue learning devices, followed by the birth of the first industrial robot, \emph{Unimate}, in the 1960s.
Recently, the unprecedented success of large language models (LLMs) has accelerated the pursuit of AGI, which in turn has revitalized the focus on embodiment as an essential pathway to human-level cognition.
This pursuit emphasizes the necessity for AI to ground its cognitive capabilities in the physical world through dynamic environmental interaction.
Such entities, possessing the dual attributes of carrier and interaction, are termed embodied agents.
From primitive mechanical prototypes to today's advanced bionic robots, they are becoming the core link connecting the digital and physical worlds for future human-machine collaboration.

\subsection{Concept and Scope Definition}
\label{Concept and Scope Definition}
To clearly delineate the boundaries between different types of agents, we categorise them into two broad classes: disembodied agents and embodied agents. 
This classification aligns with the presentation in \cref{Comparison of Research}.

\textbf{Disembodied agents,} contrasting with embodied agents, refer to intelligent entities that accomplish tasks within virtual spaces solely through digital signals or algorithmic programmes, without reliance on physical entities.
Their interactions are confined to data input and output, such as text-based conversations at the data layer and model reasoning, exemplified by purely conversational chatbots like web AI.
The core characteristic of such agents is ``disembodiment", wherein their decision-making logic bears no direct connection to physical operations in the material world.
It should be noted, however, that non-embodied agents are not entirely disconnected from the physical world. 
Their interaction with the physical realm can be achieved through intermediary vehicles, though such interaction does not involve direct physical manipulation.

\textbf{Embodied agents,} conversely, denote intelligent systems requiring physical carriers to engage in perception-action interactions with real environments.
It is noteworthy that some literature, e.g., \cite{Embodied_Agent_Modeling_arXiv}, classifies virtual-physical mapping entities incorporating digital twin technology, i.e., embodied virtual agents, within this category.   
However, this article specifically limits its scope to embodied agents that utilise real physical entities (such as industrial robotic arms or unmanned vehicles) as carriers. 
Their interaction processes must directly engage with the real physical environment, rather than relying on virtual simulations or digital twin models.
Furthermore, it is essential to distinguish embodied agents from pre-programmed mechanical devices. 
Despite possessing physical form, such devices can only execute fixed actions according to predetermined programmes, lacking autonomous interaction and environmental adaptation capabilities.

\subsection{Embodiment Value}
\label{Embodiment Value}
The embodiment of agents enables AI systems to interact directly with the physical world, thereby allowing them to assist humans in a wide range of tasks, ranging from household chores to industrial manufacturing and even military operations.
On the other hand, from the perspective of human cognitive psychology, embodiment holds equally significant value \cite{Embodiment_Value}. 
Research indicates that the embodiment of social agents markedly enhances social presence. 
When learning materials are presented by embodied virtual instructors, students demonstrate significantly higher engagement levels and knowledge retention rates compared to non-embodied formats. Concurrently, the concrete form and natural social behaviour of virtual agents effectively reduce psychological defences among users, thereby enhancing trust in interactions.
{Beyond social presence, embodiment empowers agents to learn physical laws (e.g., causality), a capability inaccessible to purely digital AI. 
Ultimately, it catalyzes the evolution from foundation AI to human-level embodied AGI, transforming agents from passive data processors into active environmental masters.}

\begin{figure*}[t]
	\centering
	\includegraphics[width=0.85\textwidth]{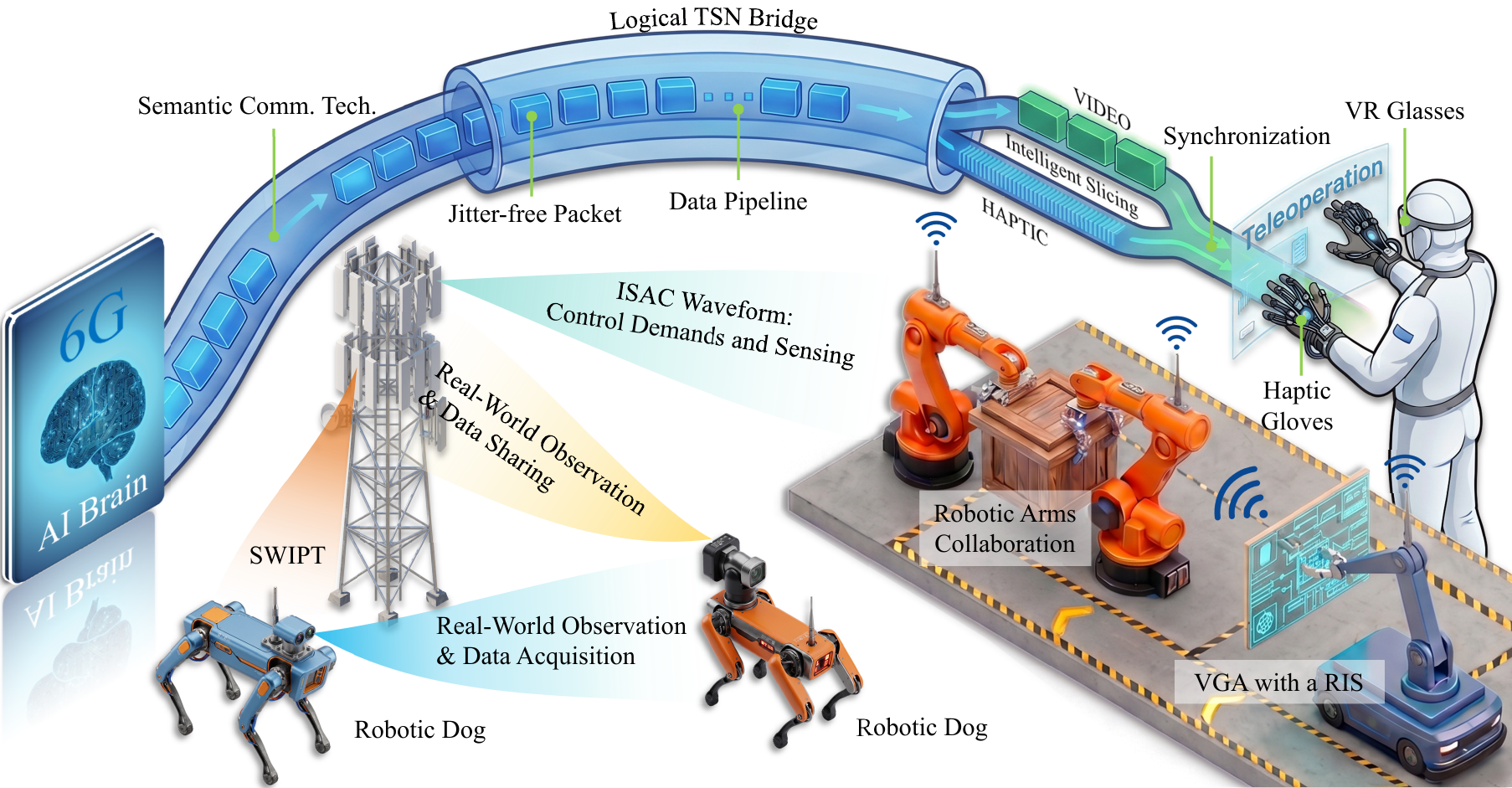}
	\caption{\label{Agent-6G} 
	The symbiotic triad: Humans, embodied agents, and 6G networks.
	}
\end{figure*}

\subsection{Challenges and Chances}
\label{Challenges and Chances}
In stark contrast to relatively static traditional Internet of Things (IoT) devices, a characteristic of embodied agents is that their autonomous actions directly impinge upon and alter the physical world. 
However, despite significant breakthroughs in robotics, the current application boundaries of embodied agents remain constrained by inherent limitations in communication capabilities.
For instance, robotic arms in industrial settings predominantly rely on localised programming for closed-loop control, capable only of executing predefined tasks at fixed workstations. 
Similarly, the mobility and decision-making of service robots are often confined to small areas covered by local area networks, making it difficult to access wide area networks for remote dynamic scheduling. 

Although 5G has made significant strides in supporting machine-to-machine communication, its optimisation efforts remain predominantly focused on passive information transmission rather than the active environmental interaction required by embodied agents.
As the vision for 6G technology unfolds, its revolutionary technological breakthroughs present an unprecedented historical opportunity to meet the demanding communication requirements of embodied agents. 
The immense potential of this opportunity has garnered broad consensus within the industry: leading global organisations including the GSMA, next G alliance, and China's IMT-2030 (6G) promotion group have explicitly identified robots as one of 6G's pivotal application scenarios \cite{AI_6G_VTM}.





\section{Symbiosis Between Embodied Agent and 6G}
\label{Symbiosis Between Embodied Agent and 6G}



The evolution of 6G extends far beyond achieving orders-of-magnitude leaps in key performance indicators (KPIs) relative to 5G.
Its deeper significance lies in driving profound expansion of application scenarios. 
For embodied agents, this signifies a transformation of network connectivity from auxiliary transmission to comprehensive empowerment.
This shift also aligns with the vision of Industry 5.0 \cite{URLLC_6G_Industry5.0_arXiv}: embodied agents are no longer cold, mechanical executors, but partners capable of understanding human intent and coexisting harmoniously with humans within physical spaces.  
Against this backdrop, as shown in \cref{Agent-6G}, 6G establishes a symbiotic collaborative relationship with embodied agents and humans.
6G serves as the digital nervous system for embodied agents, while these embodied agents integrate into 6G, forming its physical extension in the real world.

\subsection{6G Empowering Embodied Agent System}
\label{6G Empowering Embodied Agent System}

For embodied agents undertaking dynamic tasks, particularly in scenarios involving high-precision teleoperation, the demands on communication latency and reliability are exceptionally stringent.
Whilst 5G ushered in the era of ultra-reliable and low latency communication (URLLC), its capabilities approach saturation when confronted with the stringent demands of next-generation embodied agents. 
6G evolves this paradigm into enhanced URLLC (eURLLC), aiming for revolutionary KPI improvements: air interface latency is projected to decrease from millisecond levels to below 0.1 ms, while reliability targets rise from 99.999\% to 99.99999\% \cite{6G_Survey_COMST}.

Moreover, the sub-scenarios evolved from URLLC can accommodate the diverse operational requirements of embodied agents.
\textbf{1) Massive ultra-reliable low-latency communication (mURLLC):} 
Designed for dense environments such as smart factories, it enables high-reliability, low-latency coordination among vast numbers of embodied agents, ensuring efficient execution of large-scale clustered tasks.
\textbf{2) Mobile broadband reliable low latency communication (MBRLLC):} 
Numerous embodied agent tasks, such as visual servoing, require high-definition video streams (demanding high bandwidth) and control signalling (demanding low latency) to be transmitted over the same link, which often presents a fundamental trade-off in 5G. 
6G MBRLLC addresses this challenge by integrating these two capabilities, thereby enabling more innovative embodied intelligence applications.
For instance, immersive remote operation combined with extended reality (XR) technology enables operators to ``be present" at the site for high-fidelity interaction.

Additionally, for the control stability of physical systems, minimising jitter is often more critical than reducing average latency. 
Thanks to advances in deterministic networking, 6G is projected to control jitter at the microsecond level \cite{6G_Survey_COMST}. 
A mechanism called logical time-sensitive networking (TSN) bridge \cite{TSN_Survey_COMST} masks underlying wireless fluctuations {by implementing deterministic mechanisms (such as precise time synchronization and time-aware shaping) and utilizing hold-and-forward buffers to absorb wireless jitter.} This allows the 6G system to function as a transparent black box that simulates the deterministic physical properties of Ethernet cables.
This determinism is vital for maintaining the stability of closed-loop control.
Robotic control systems often demand cycle times of 1 ms with jitter strictly below 1 µs.
Even if just 1\% of commands arrive delayed by 10 ms due to network fluctuations, it may lead to execution deviations, system instability (such as a robot falling over), or severe safety risks to human collaborators.

However, wireless certainty is fundamentally constrained by dynamic physical environments. 
6G networks should not merely serve as passive transmission conduits but evolve into intelligent entities possessing situational awareness.
Leveraging integrated sensing and communication (ISAC) technologies, 6G networks can see physical environmental changes, often induced by an embodied agent's own movement (e.g., a robotic arm blocking a signal path). 
By integrating agentic AI, 6G networks can understand the stochastic behavior of the wireless channel and disclose pertinent information to the embodied agent, enabling a shift from reactive control to predictive adaptation.
For instance, milliseconds before a potential link outage, the network can proactively establish an alternative reachable path via reconfigurable intelligent surfaces (RIS), ensuring seamless resilience.

Furthermore, time synchronisation is another critical metric. 
At the collaborative robot level, such as in dual-arm cooperative heavy lifting, even millisecond-level asynchrony can generate destructive mechanical stress, leading to task failure or hardware damage.
More critically, this synchronisation requirement is further amplified within closed-loop control systems featuring human-in-the-loop operations. 
A significant psychophysical constraint arises here: humans exhibit high tolerance for visual latency (approximately 100 ms) yet extreme sensitivity to tactile feedback (10 ms). 
In tasks like telesurgery or bomb disposal, sensory conflict can trigger operational oscillations or even motion sickness if these two modalities fail to align precisely. 
As video streams involve substantial data volumes, semantic communication techniques can be employed to compress video data, thereby significantly reducing the arrival time difference between video streams and high frequencies haptic signals at the transmission level.
Besides, 6G networks introduce multimodal collaborative scheduling mechanisms \cite{URLLC_6G_Industry5.0_arXiv}. 
Leveraging native AI, the network identifies video, haptic, and control streams belonging to the same embodied session, enforcing constraints on their latency skew. 
Where necessary, the network may even implement deliberate delay strategies. 
For instance, faster haptic signals may be micro-buffered to ensure visual and haptic information arrive synchronously within the operator's perceptual threshold. 
{Although beneficial for sensory alignment, this intentional delay increases the overall latency, which negatively impacts the operator's control experience. 
Therefore, dynamic balance between synchronization and latency is worthy of consideration.}

{Apart from the aforementioned performance metrics, 6G can also enhance the endurance of mobile robots. 
By combining semantic communication techniques capable of reducing data overhead with simultaneous wireless information and power transfer (SWIPT) techniques for energy harvesting, 6G helps energy-constrained robots extend their working time.}


\subsection{Embodied Agents Reciprocate 6G Network}
\label{Embodied Agents Reciprocate 6G Network}

The success of 6G hinges not only on network capabilities but, more crucially, on the development of 6G terminals.
Embodied agents are no longer mere consumers of communication networks.
They are evolving into integral components of the network infrastructure itself. 
Leveraging their mobility, these embodied agents can assist in extending signal coverage. 
For instance, by carrying RIS, these agents can effectively eliminate coverage blind spots inaccessible to fixed infrastructure. 
Furthermore, equipped with rich sensor arrays, embodied agents can function as sensing probes for the 6G AI brain, capable of gathering data such as channel state information (CSI) and environmental semantics to inform the network's overall intelligent optimisation.
{At a deeper level, embodied agents reciprocate by sharing their physical understanding---acquired through real-world interactions---with the 6G network. Instead of merely transmitting raw data, agents provide causal predictions, such as anticipating the movement trajectories of obstacles. This endows the network with causal reasoning, enabling it to proactively predict dynamic blockages and optimize resource allocation in advance, ultimately driving the evolution toward AGI-native wireless systems \cite{AGI_6G_IEEE_Proceedings}.}

\section{Communication System for Human-Robot Remote Interaction}
\label{Communication System for Human-Robot Remote Interaction}

\subsection{Architecture Design}
\label{Architecture Design}
\begin{figure}[t]
	\centering
	\includegraphics[width=0.49\textwidth]{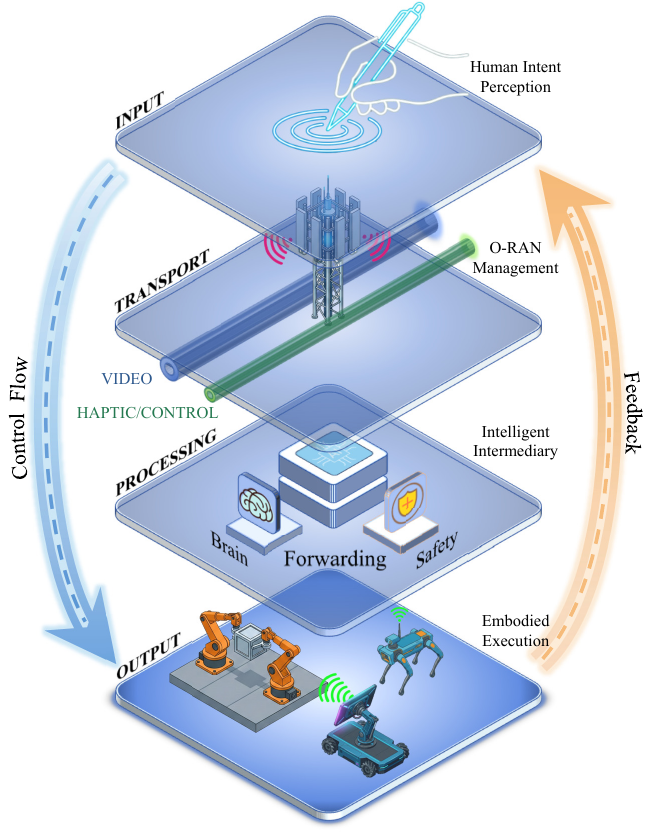}
	\caption{\label{Architecture}
	Hierarchical architecture for 6G-enabled human-robot remote interaction.
	This architecture supports flexible control strategies (encompassing open-loop, closed-loop, and dual-loop switching), utilises the RIC within O-RAN to manage multi-modal data transmission, and facilitates customisation incorporating features such as LLM-enhanced intelligence and secure authentication.
	Furthermore, the architecture is compatible with diverse heterogeneous embodied agents, enabling plug-and-play functionality. 
	}
\end{figure}

In the context of Industry 5.0, we propose a hierarchical architecture for remote human-robot collaboration. 
As illustrated in \cref{Architecture}, this architecture leverages open radio access network (O-RAN) as the transport backbone and introduces an intelligent intermediary platform as the system's cognitive hub.
This design transcends traditional point-to-point teleoperation by decoupling control logic from physical connectivity, enabling scalable, secure, and intelligent collaboration.

\subsubsection{Human-intent perception layer}
\label{Human-intent perception layer}
This architecture commences with the fused agent, a symbiotic entity integrating human cognition with a cognitive conversion device, which employs intelligent sensors to capture human operational intent, e.g., grasping a target object. 
Leveraging LLMs or domain-specific knowledge bases, it translates abstract intent into control commands recognisable by embodied agents, thereby achieving the digitalisation of operational intent.

\subsubsection{O-RAN layer}
\label{O-RAN layer}
This layer connects operators with remote agents. 
As a key candidate for 6G, O-RAN offers a flexible, agile and decoupled network architecture \cite{ORAN_NM}. 
Through the RAN intelligent controller (RIC) within O-RAN, this layer actively manages targeted network slicing for distinct data modalities.
For instance, URLLC slices are prioritized for high-priority haptic/control signals to ensure sub-millisecond jitter while enhanced mobile broadband (eMBB) slices accommodate high-bandwidth video streams. 
{This slicing mechanism prevents network congestion, safeguarding system stability and mitigating severe east-west traffic contention when scaling to large-scale agent clusters.}

\subsubsection{Intelligent intermediary layer}	
\label{Intelligent intermediary layer}
Functioning akin to a cognitive neural hub, this layer supports the construction of personalised functionalities, such as  forwarding commands, implementing safety guardrailing to filter non-compliant or hazardous instructions, or integrating LLMs to comprehend complex tasks.
{To mitigate potential LLM overhead, this layer can adopt a decoupled deployment: cloud- or edge-based LLMs asynchronously parse complex intents, while the local platform directly handles high-frequency kinematic control.}
{Meanwhile, commands generated by LLMs can be pre-verified via local digital twin simulations before physical execution. Furthermore, live video and digital twin feedback keeps the human in the loop, enabling operators to quickly detect and correct any semantic misinterpretations.}

\subsubsection{Embodiment layer}
\label{Embodiment layer}
As the final layer, embodied agents respond to upper-layer instructions and feed back diverse data. 
This includes not only critical sensor readings but also self-discovered insights, such as local anomalies or world model updates, ensuring the system remains deeply synchronized with the physical world.

\subsection{Prototype Implementation}
\label{Prototype Implementation}
\begin{figure*}[t]
	\centering
	\includegraphics[width=0.85\textwidth]{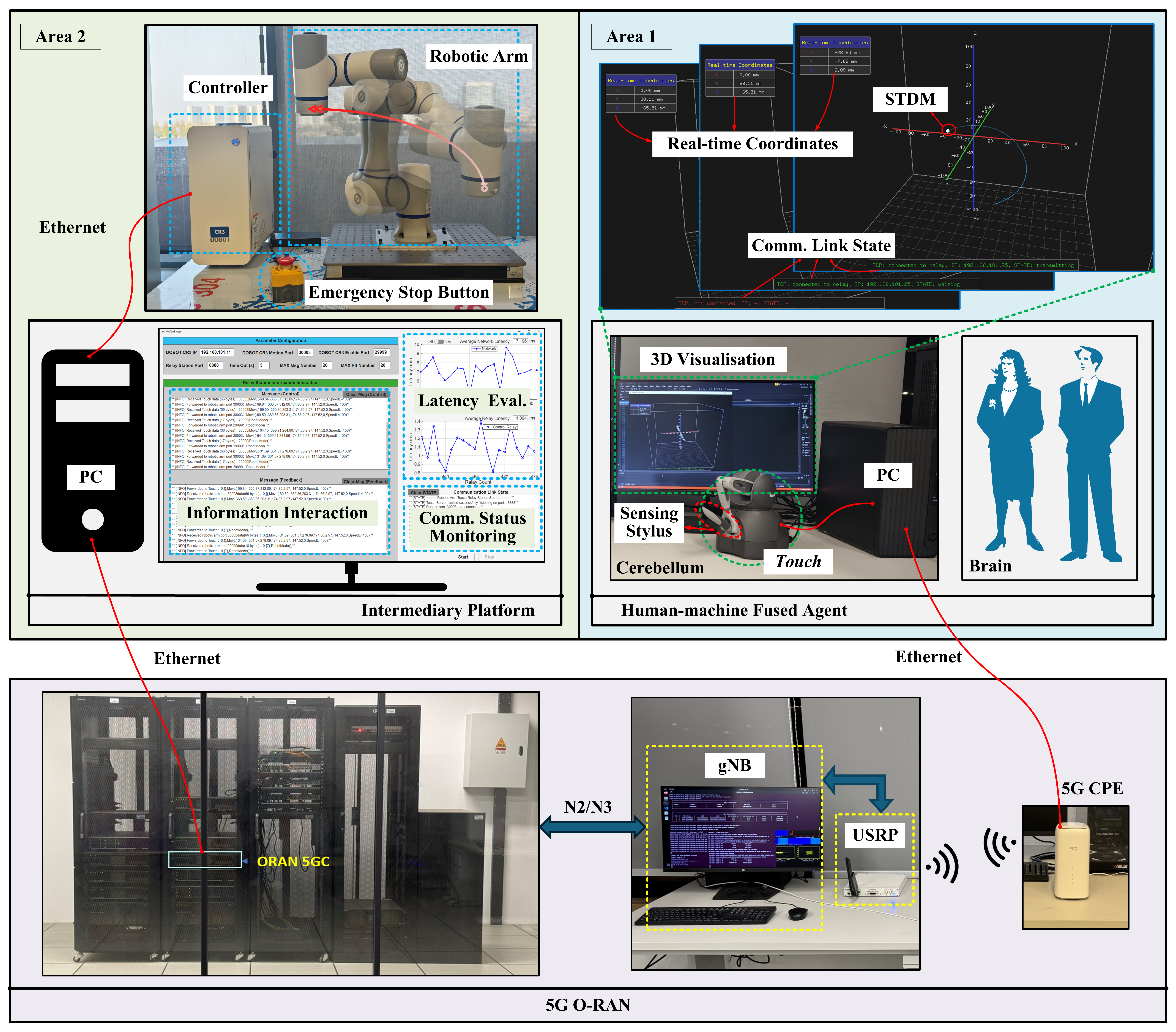}
	\caption{\label{system} 
	Prototype for human-robotic arm remote interaction.  
	A user at area 1 wants to control a robotic arm at area 2 via the \emph{Touch} device. 
	Transmitted to the O-RAN network through Ethernet, the commands are forwarded to the intermediary platform for evaluation, and compliant ones are sent to the robotic arm. 
	A visual interface is developed with stylus tip digital mapping (STDM), real-time control trajectories, and its three-dimensional (3D) coordinates.
	}
\end{figure*}

To provide a reference framework and practical model for embodied agent communication research, this article constructs an end-to-end human-robot remote collaboration prototype system, as shown in \cref{system}.

\subsubsection{Overall System}
\label{Overall System}

This system employs a remote real-time bidirectional interaction mode, in which a robotic arm interacts with a human-machine fused agent.
Among them, the fused agent follows the specific operational logic as follows.
Humans serve as the decision-making hub that provides movement intent (i.e., the brain), while performing spatial displacement movements via a sensing stylus on a haptic interaction device called \emph{Touch}.
Acting as the cerebellum, the \emph{Touch} device functions as a perception-action intermediary, capturing real-time spatial coordinate data from the stylus tip, translating human movement intent into standardised control commands, and driving synchronous motion in the remote robotic body. 
More specifically, the system is composed of four core modules: a \emph{Touch}, a robotic arm, an intermediary platform, and a 5G O-RAN.
Their specific details are outlined below.

\textbf{Haptic device:} \emph{Touch} is an electric force feedback haptic interaction device, designed primarily to deliver immersive human-machine interaction. 
By applying precise force feedback to the user's hand, it delivers a lifelike tactile experience when simulating the sensation of touching virtual objects.
Its built-in multi-dimensional sensors provide real-time detection of the sensing stylus's X/Y/Z-axis positioning, alongside pitch/roll/yaw orientation data from the stylus's universal joint, delivering precise data support for interactions. 
The device connects to a personal computer (PC) via a USB interface and supports development in programming languages such as C++, allowing customisation of functions based on its hardware capabilities. 
Within the human-robotic arm interaction system, it serves as a core input device, translating the user's control intentions into precise data to support control of the remote robotic arm.

\textbf{Robotic arm:} The core of this robotic arm system comprises a control cabinet, a six-axis robotic arm unit, and an emergency stop button. 
Based on the TCP protocol, the robotic arm system can establish a communication connection with the intermediary platform via WiFi or Ethernet, thereby supporting secondary development.
Upon receiving a motion command, the robotic arm autonomously calculates the rotational angles of each joint through its built-in kinematic algorithms, then utilises its attitude sensors to verify its own position.
It is worth noting that, in order to achieve correct control of the robotic arm via the \emph{Touch}, the spatial coordinates of the \emph{Touch} device must be aligned with those of the robotic arm.

\textbf{Intermediary platform:} 
Using MATLAB, we developed the intermediary platform, capable of performing compliance verification and targeted forwarding on the control commands issued by the user terminal and the feedback data from the robotic arm.
Furthermore, leveraging this platform enables the quantitative assessment of two critical performance metrics: network latency within communication links and the time consumption of the platform's own data forwarding processes.
Additionally, a window is provided for monitoring communication connections.

\textbf{5G O-RAN:}
To support remote communication for the agents, we constructed a 5G O-RAN network.
The \emph{Touch} device establishes a connection with the 5G customer premises equipment (CPE) via Ethernet. 
The CPE further communicates with the gNodeB (gNB) through a 5G wireless network. 
The gNB, implemented based on OpenAirInterface (OAI), comprises three functional modules: the central unit (CU), distributed unit (DU), and radio unit (RU). 
The radio frequency (RF) layer functionality of the RU is implemented by a universal software radio peripheral (USRP) device equipped with two rubber rod antennas. 
Communication between the CU and the 5G core network (5GC) is established via N2 and N3 interfaces. 
The 5GC, also based on OAI, runs in Docker containers on rack mounted servers in our data center.
Operating at 3.3 GHz with a 40 MHz bandwidth, the access network employs in time division multiplexing mode, implementing a complete 5G protocol stack.

\subsubsection{Control Workflow}
\label{Control Workflow}

When the operator moves the stylus tip of the \emph{Touch} device, the device can capture its real-time coordinates.
Simultaneously, if the operator presses the data transmission button on the stylus, these real-time coordinates are converted into motion commands recognisable by the robotic arm. 
These control signals are then transmitted via the 5G O-RAN network to the intermediary platform. 
Once the transmission button is released, the conversion and transmission of commands cease.
Upon receiving and validating the command information, the intermediary platform forwards it to the corresponding port of the robotic arm. 
The robotic arm then parses the command content into execution signals for joint movement and posture adjustment, ultimately completing the physical execution phase of human-robot interaction.

\subsubsection{Feedback Mechanism}
\label{Feedback Mechanism}
To ensure control stability, we adopted a closed-loop control approach, necessitating timely feedback of the robotic arm's status.
Each time the robotic arm executes one command, it relays feedback regarding the command's execution status to the intermediary platform. 
Upon receiving this feedback, the intermediary platform promptly transmits it to the \emph{Touch} device terminal, enabling the operator to monitor the robotic arm's operational status in real time, such as whether an alarm has been triggered or if the command has been successfully executed.

Within this system, different types of commands correspond to distinct feedback information. 
For routine commands such as motion control, the robotic arm automatically provides feedback on the specific execution status of the command upon receipt. 
For query commands, the robotic arm returns the corresponding query results. 
A typical application scenario involves operators periodically querying whether the robotic arm is in an alarm state to ensure smooth remote operation. 
If the coordinates received by the robotic arm exceed its kinematically feasible range, an alarm mechanism is immediately triggered. 

\subsubsection{Visualisation Interface Development}
\label{Visualisation Interface Development}
To enhance the intuitive operation and interactive experience for \emph{Touch} operator, we developed a 3D visualisation interface. 

\textbf{3D space rendering:} The interface incorporates a 3D coordinate axis with numerical graduations, dynamically mapping the current spatial position of the stylus tip via a real-time cursor.
A short-term motion trail generated during cursor movement visually represents the trajectory, enhancing the operator's spatial awareness and thereby increasing confidence in system control.
Meanwhile, one data table synchronously displays the 3D coordinates from the stylus tip.


\textbf{Communication state monitoring:} A connection status bar provides real-time updates on the connection status of the communication link.
The current system employs TCP protocol for communication between the \emph{Touch} device and the intermediary platform, and between the intermediary platform and the robotic arm.
It is worth noting that the communication protocol between the intermediary platform and the \emph{Touch} device can be flexibly customised, such as switching to UDP for lower latency.
Furthermore, if LLM agents are integrated, existing agent collaboration protocols proposed within the research community, e.g., A2A and ACP, could be utilised.


\subsubsection{Transmission  Optimisation}
\label{Transmission  Optimisation}
This system performs real-time traffic statistics on the 5G O-RAN network, and leverages adaptive modulation capabilities to balance transmission reliability and efficiency.
When channel quality deteriorates, the network automatically switches to lower-order modulation schemes, prioritising data transmission reliability. 
When channel conditions improve, the network increases the modulation order, simultaneously enhancing transmission rate. 

\subsubsection{System Performance}
\begin{figure}[t]
	\centering
	\includegraphics[width=0.49\textwidth]{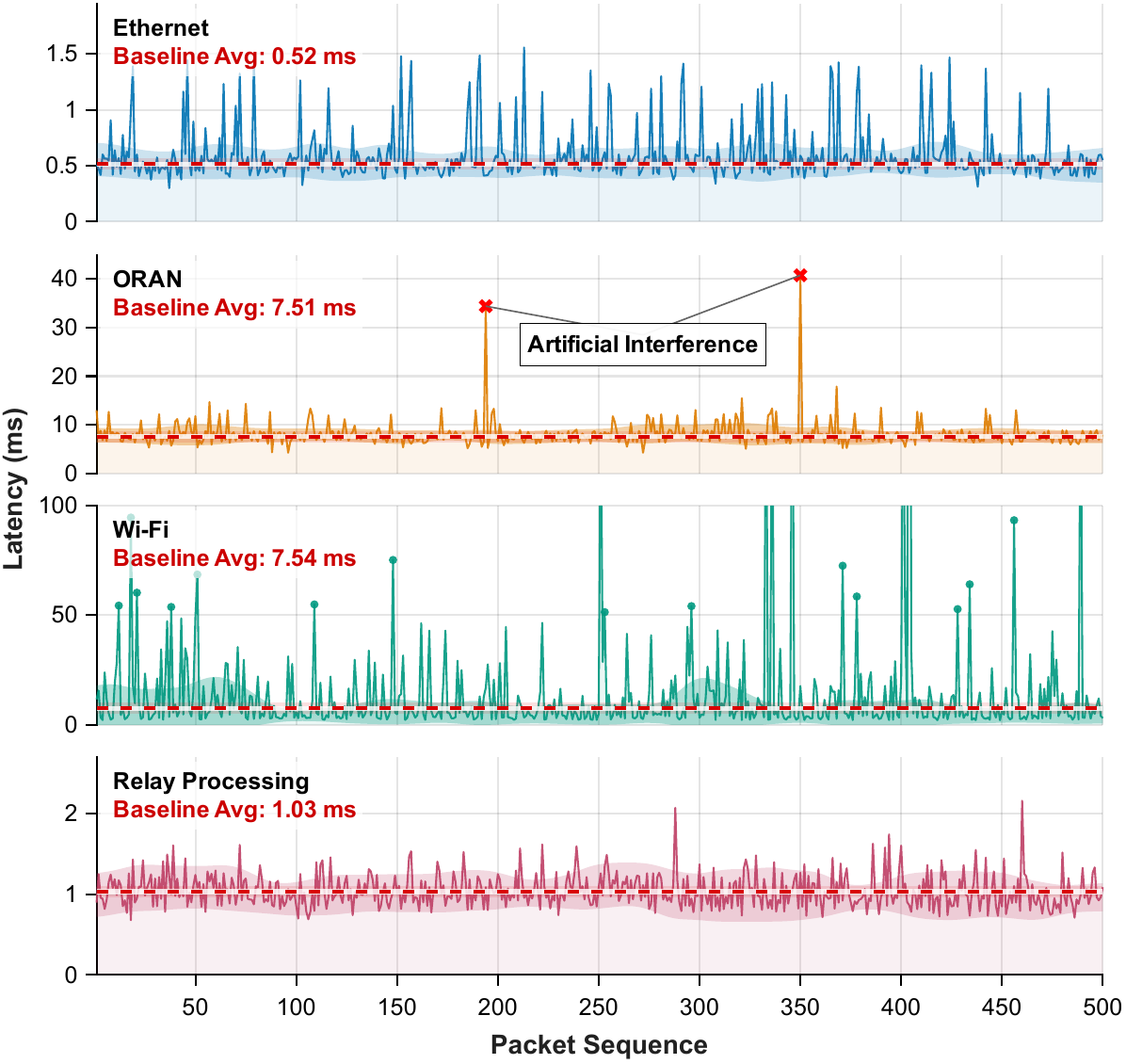}
	\caption{\label{Latency_results}
	Experimental comparison of network latency and jitter.
	{All network tests employ small packets ($\approx$  32 bytes, TCP protocol), with transmission latency estimated as half of the round-trip time.}
	Ethernet demonstrates the highest stability with minimal delay.
	O-RAN performance is generally stable, with the exception of two distinct spikes induced by {intentional line-of-sight blockage}. 
	While Wi-Fi shares a comparable baseline average with O-RAN, it exhibits significant jitter due to spectrum contention in the shared laboratory setting. 
	The bottom panel quantifies the relay processing overhead, including instruction compliance verification and forwarding.
	{Note that these results are tied to hardware and software performance.}
	}
\end{figure}

\label{System Performance}
\textbf{1) Latency:} Utilising the intermediary platform, we tested the transmission latency of networks alongside the processing latency associated with the intermediary platform's forwarding operations. 
As shown in \cref{Latency_results}, laboratory testing demonstrated that the average transmission latency of the 5G O-RAN was below 8 milliseconds, whilst the average latency for the intermediary platform forwarding control commands was under 2 milliseconds. 
Furthermore, as the intermediary platform and the robotic arm reside on the same subnet, the transmission latency between them was measured to be less than 1 ms.
This performance fully meets the latency requirements for real-time remote human-robot control, satisfying, for instance, the 200 ms latency threshold required for remote surgery.
{Looking forward, key 6G technologies, such as deterministic networking, AI-enabled slicing, and ISAC, can further enhance system stability by decreasing jitter and bypassing potential air interface blockage.}

\textbf{2) Actual control effect:}  Within this system, the control commands transmitted from the control terminal to the robotic arm exhibit characteristics of high frequency and small data volume.
Due to hardware limitations such as the robotic arm's joint response speed and motor torque, its command execution rate falls below the data reception rate. 
This results in the phenomenon of lag in robotic arm movements during high-speed teleoperation.
For instance, when \emph{Touch} completes two full rotations, the robotic arm responds with only half a rotation. 
To mitigate this lag, a passive approach is to downsample the control commands, which however compromises trajectory reproduction.
{To resolve the trade-off between responsiveness and trajectory accuracy, predictive algorithms provide a task-adaptive solution. 
For operations prioritizing high responsiveness, algorithms like model predictive control (MPC) dynamically discard outdated intermediate data and forecast forward trajectories based on the latest inputs. This allows the robotic arm to striv to keep up with the latest commands based on its hardware capability. 
Conversely, for tasks prioritizing high precision (e.g., writing), by anticipating delayed commands, predictive algorithms can absorb network jitter and ensure the robot fluidly executes the full geometric path, thereby preventing mechanical jerkiness.}


\subsubsection{Sensory Extension}
\label{Sensory Extension}
With the successful establishment of this system's control and feedback loops, further functional expansion has become feasible. 
By equipping the end-effector of the robotic arm with sensing devices, the system's perceptual capabilities can be significantly enhanced and its functional scope broadened. 
Two typical examples are the integration of cameras and pressure sensors.
{Specifically, cameras provide real-time visual information, enabling environmental awareness for precise and immersive remote control. Meanwhile, force feedback from pressure sensors provides a sense of telepresence, allowing the operator to feel physical interactions.}

\section{Conclusion}
\label{Conclusion}
In this article, we reviewed the brief history, concepts, and value of embodied agents, emphasising their hardware-based physicality to distinguish them from disembodied agents.
Then, we explored the emerging symbiotic relationship between embodied agents and 6G: 6G serves as the digital nervous system, while agents function as the physical extension of the network. 
Against the backdrop of Industry 5.0, we proposed an end-to-end communication architecture centred on O-RAN and intelligent intermediary platforms for human-robot remote interaction. 
Further, we developed a prototype system implementing closed-loop control and validated the effectiveness of this architecture.
Looking ahead, this prototype holds promise for further realising interaction between embodied and disembodied intelligence.

\balance
\bibliography{reference/AGENT_reference}

\end{document}